\documentclass[letterpaper, 10 pt, conference]{ieeeconf}  

\IEEEoverridecommandlockouts                              
\usepackage{amsmath}
\usepackage{amsfonts}
\usepackage{amssymb}
\usepackage{graphicx}
\usepackage{booktabs}
\usepackage{subcaption}
\usepackage[font={small}]{caption}
\usepackage{xcolor}
\let\labelindent\relax
\usepackage{algorithm,algpseudocode,enumitem}
\usepackage{amsmath}
\usepackage{cite}
\usepackage{hyperref}

\title{\LARGE \bf
Emergent Hand Morphology and Control\\from Optimizing Robust Grasps of Diverse Objects
}

\author{Xinlei Pan$^{1,2}$, Animesh Garg$^{1,3}$, Animashree Anandkumar$^{1,4}$, Yuke Zhu$^{1,5}$
    \thanks{ $^{1}$NVIDIA, $^{2}$University of California, Berkeley; $^{3}$University of Toronto, Vector Institute; $^{4}$Caltech; $^{5}$The University of Texas at Austin}
}

\begin{document}


\maketitle
\thispagestyle{empty}
\pagestyle{empty}

\begin{abstract}
Evolution in nature illustrates that the creatures' biological structure and their sensorimotor skills adapt to the environmental changes for survival. Likewise, the ability to morph and acquire new skills can facilitate an embodied agent to solve tasks of varying complexities. In this work, we introduce a data-driven approach where effective hand designs naturally emerge for the purpose of grasping diverse objects. Jointly optimizing morphology and control imposes  computational challenges since it requires constant evaluation of a black-box function that measures the performance of a combination of embodiment and behavior. We develop a novel Bayesian Optimization algorithm that efficiently co-designs the morphology and grasping skills jointly through  learned latent-space representations. We design the grasping tasks based on a taxonomy of three human grasp types: power grasp, pinch grasp, and lateral grasp. Through experimentation and comparative study, we demonstrate the effectiveness of our approach in discovering robust and cost-efficient hand morphologies for grasping novel objects.
\url{https://xinleipan.github.io/emergent_morphology/}
\end{abstract}

\section{Introduction}


Natural life provides ample evidence of a history of the body, the mind, and the environment all evolving together in cooperation with, and in response to each other~\cite{mautner2000evolving}. An agent not only learns a policy with immutable morphology, but the morphology itself adapts (slowly) to the distribution of tasks it is exposed to. This shows up in many instances such the co-emergence of adducted thumb along with larger cognitive processing ability in humans. As humans developed rare abilities, such as tool use, throwing, and clubbing~\cite{young2003evolution}, both the hand and the ability to use it emerged. Likewise, for a robot to solve a task, say manipulating objects, the interplay between the robot embodiment (hardware) and its control algorithm (software) has an integral role in  task success. 

Meanwhile, recent advances in learning for robotics have yielded studies on data-driven techniques for robot grasping~\cite{mahler2017dex,mahler2017dex2}. Most of them have focused on grasp planning and selection~\cite{Shao2019UniGraspLA}, i.e., the control algorithm of grasping, with fixed and often simple parallel-jaw grippers. 
In contrast, customizing the robot embodiment to a particular task simplifies the computational burden of decision making, a phenomenon referred to as \emph{morphological computation} in robot control~\cite{pfeifer2009morphological}. For the task of grasping with different modes, robot hands with sophisticated mechanical mechanisms, such as passive compliance ~\cite{backus2016adaptive,deimel2016novel}, suction end-effectors~\cite{mahler2017dex,eppner2017lessons}, and actuated rolling fingertips~\cite{yuan2020design} have been carefully engineered to simplify grasp planning problems and improve performance.
Though the data-driven approaches have given rise to more robust grasps on various objects from raw sensory data, their effectiveness has been fundamentally handicapped due  to the neglect of the impact of the robot embodiment, where failure modes are often attributed to the limitation of the gripper designs.

\begin{figure}[t!]
    \centering
    \includegraphics[width=0.95\linewidth]{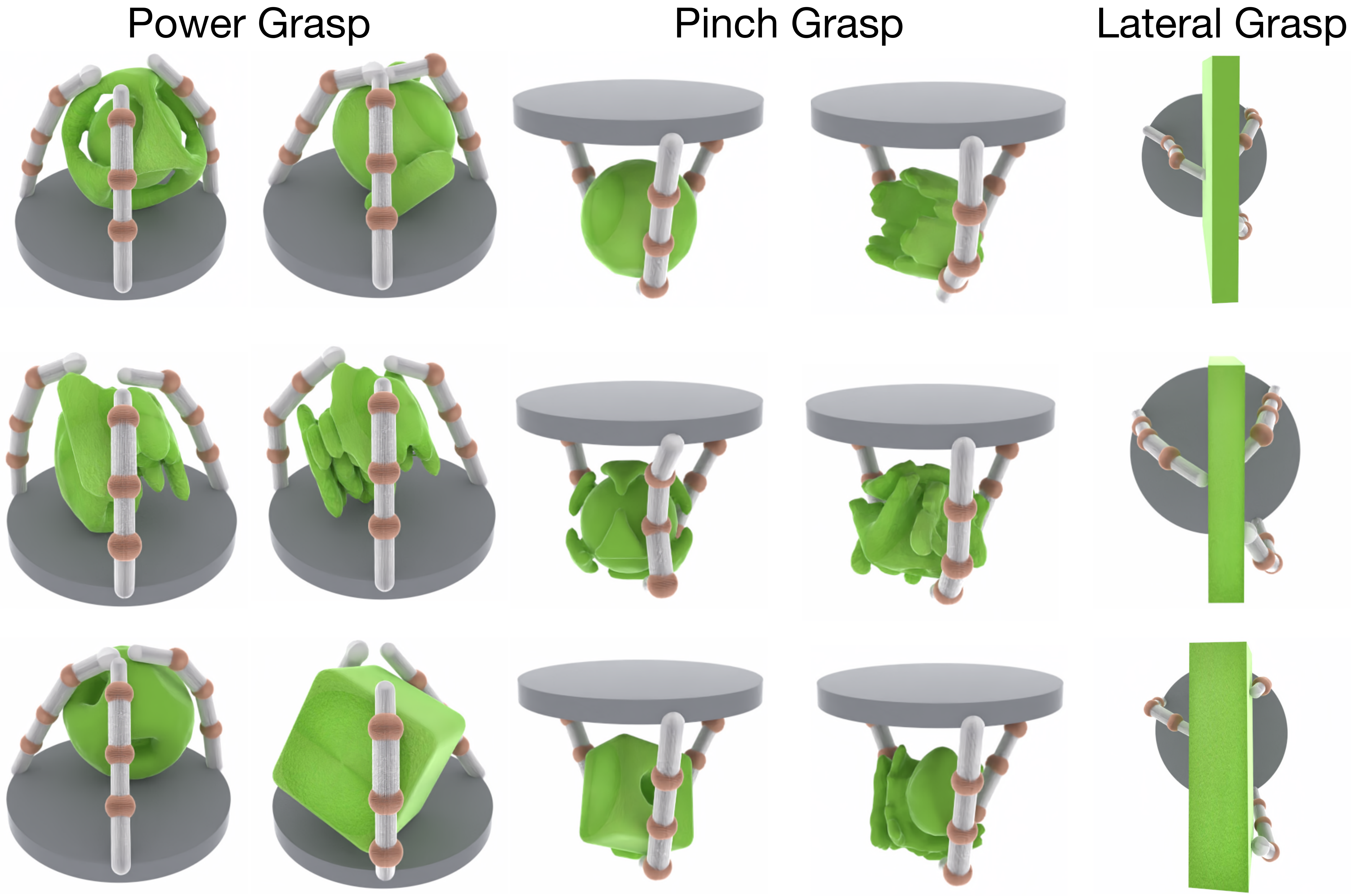}
    \caption{
     \textbf{Emergent hand morphology and behavior.} We optimize hand design and control on grasping a diverse set of objects, aiming at maximizing the success rate and minimizing the hand complexity. Here are examples of the emergent hand design and grasping behavior from our model in three types of grasps. Our optimized hand is both effective and cost-efficient in its design.}
    \label{fig:optimized-hand}
    \vspace{-10pt}
\end{figure}

In this work, we introduce a data-driven approach to jointly optimizing robot embodiment and behavior for object grasping, with the aim of maximizing grasp performance. While optimizing for a specific task may lead to ill-formed hand designs that specialize at the task but fails on others, we demonstrate that robust and cost-efficient hand designs emerge from grasp optimization on a diverse of objects in three distinct grasp types (power, precision, and lateral), as classified under the taxonomy of human grasps~\cite{feix2015grasp}. In Figure~\ref{fig:optimized-hand} we show example hands optimized considering the performance and cost all together by optimizing the design on a diverse set of objects.




\begin{figure}[t!]
    \centering
    \includegraphics[width=1.\linewidth]{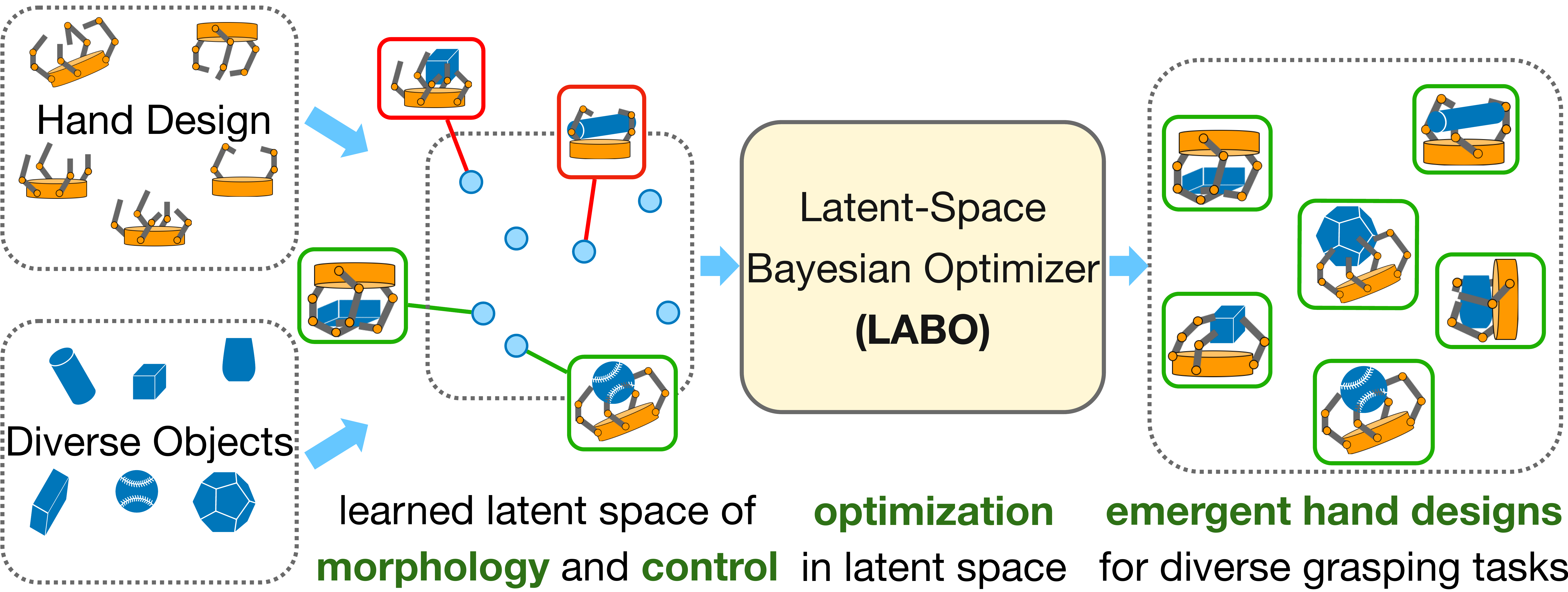}
    \caption{\textbf{High-level overview of our optimization framework.} We jointly optimize the hand design and the control policy to solve diverse object grasping tasks. With representation learning of the design parameters in a latent space, our Bayesian optimizer (LABO) finds a robust and cost-efficient hand design for various grasp types.}
    \label{fig:goal}
    \vspace{-10pt}
\end{figure}

The key challenge of this problem resides in the high dimensionality of the optimization problem, involving a co-adaptation of the morphology and control parameters. The quality of the parameters can only be faithfully assessed through empirical grasping trials. However, the high costs incurred for conducting grasping trials, even in physics-based simulation, make it prohibitive for conventional black-box optimization techniques~\cite{hansen2003reducing,frazier2018tutorial}. Our intuition is that only a subset of design parameters have a strong impact on the grasping performance. Therefore, we can reduce the problem size by extracting these important factors through learning. 
To this end, our method learns latent representations of morphology and control parameters that  capture the importance and redundancy of design parameters and thereby facilitate knowledge transfer across similar morphologies. Figure~\ref{fig:goal} shows a high-level overview of our latent-space optimization framework for hand design.


\textbf{Summary of Contributions.} (1) We develop a novel Bayesian Optimization (BO) method that harnesses a learned latent space of morphology and control parameters for efficient co-adaptation of both. (2) We design an auxiliary loss function which predicts the performance of the morphology and control design along with variational auto-encoder (VAE) to train an encoder that cast raw parameters into a latent feature space where important factors are magnified, which accelerates the optimization process by around 100 iterations.
(3) Extensive experiments show that our approach outperforms standard black-box optimization baselines by around 20\% of success rate. We also demonstrate that the emergent design with simple but effective three-fingers hand, trained on a diverse set of 3D shapes, generalizes well to novel objects unseen during training. 

\section{Related Work}
\textbf{Optimization-Based Morphological Design.} 
The physical embodiment impacts a robot's potentials and limitations of dexterous behaviors. Carefully engineered robot designs have been shown to simplify the downstream computational problems of grasping and manipulation~\cite{pfeifer2009morphological,backus2016adaptive,deimel2016novel,eppner2017lessons,yuan2020design}. While robot designs have drawn direct inspirations from real-life organisms~\cite{seok2014design,graichen2015control,jayaram2020scaling}, it remains an immense challenge in dealing with numerous design parameters and practical constraints. Optimization-based  methods have been used for automating the co-design of morphology and control. These include gradient-free methods, such as evolutionary algorithms~\cite{leger1999automated,cheney2013unshackling,luck2020data}, and gradient-based methods, such as constrained optimization~\cite{wampler2009optimal,ha2017joint}, differentiable simulation~\cite{NIPS2019_9038,hu2019chainqueen}, and policy optimization~\cite{chen2020hardware}.
However, these methods require either great amount of samples to search in  the high-dimensional design space or manual specification of constraints. Our method efficiently co-optimizes the robot morphology and the control policy for grasping.

\vspace{1mm}
\textbf{Knowledge Sharing Across Embodiments.} Most of the existing studies for embodied agents focus on generalizing towards environmental variations for a specific agent. Sharing knowledge across agents of different embodiments has gained an increased attention~\cite{NIPS2019_9038,wang2019neural,pathak2019learning,chen2018hardware,huang2020smp}. Knowledge transfer is typically achieved by either exploiting the compositional structures of embodiment~\cite{wang2019neural,pathak2019learning,huang2020smp} or by learning shared representations of the embodiment~\cite{NIPS2019_9038,huang2020smp}. This has enabled effective transfer of robot skills~\cite{wang2019neural}, fast adaptation to test-time changes~\cite{pathak2019learning}, and efficient policy training with new morphologies~\cite{chen2018hardware}. Most relevant to ours is learning-in-the-loop optimization~\cite{NIPS2019_9038}, which learns a latent space for designing soft robots. However, they have to rely on a fully-differentiable physical simulation for evaluating and updating the embodiment.

\vspace{1mm}
\textbf{Data-Driven Grasping Methods.} Recent advances in data-driven grasping methods~\cite{mahler2017dex2,mahler2017dex,bohg2013data,lenz2015deep,levine2018learning,pinto2016supersizing} have substantially improved the performance of robotic grasping in unstructured environments from raw sensory input. To cope with the data-hungry nature of learning grasping models, especially with deep networks, simulated data has been extensively used as the training sources~\cite{fang2018tog,bousmalis2018using,viereck2017learning}. While some works have recently explored grasping and manipulation of dexterous or soft hands~\cite{andrychowicz2020learning,gupta2016learning}, the main body of the data-driven grasping literature has centered on simple and fixed end-effectors, often the parallel-jaw grippers, where the grasp prediction problem can be reduced to computing the gripper pose. Closest to our work is UniGrasp~\cite{Shao2019UniGraspLA}, which learned a gripper representation in order to transfer grasps across a set of $N$-fingered robotic hands. However, it only considers a fixed set of grippers rather than optimizing the hand morphologies.

\vspace{1mm}
\textbf{Latent Space Bayesian Optimization}. Latent space BO has been explored in chemistry and medical domains, where prior approaches have explored unsupervised learning methods, such as variational auto-encoder (VAE)~\cite{dhamala2018high,griffiths2017constrained}, to perform dimension reduction of the search space for BO. 
In contrast, we introduce auxiliary supervision of grasping success labels to expedite task-oriented representation learning, and use empirical simulated grasping trials with the latent-space Bayesian optimizer for data-driven hand design.


\section{Morphology Design with Latent Space Bayesian Optimization}


We introduce a data-driven approach to emergent morphological and policy design. Inspired by the morphological evolution of human hands~\cite{napier1962evolution} and the task-oriented optimization of robot design~\cite{van2009optimal}, we study the problem of jointly optimizing the morphology and control through grasp trials on a diverse set of objects. We examine how joint optimization on these diverse tasks could generate a robust hand morphology and control design that generalizes to unseen scenarios.

\begin{figure}[t!]
\centering
\includegraphics[width=\linewidth]{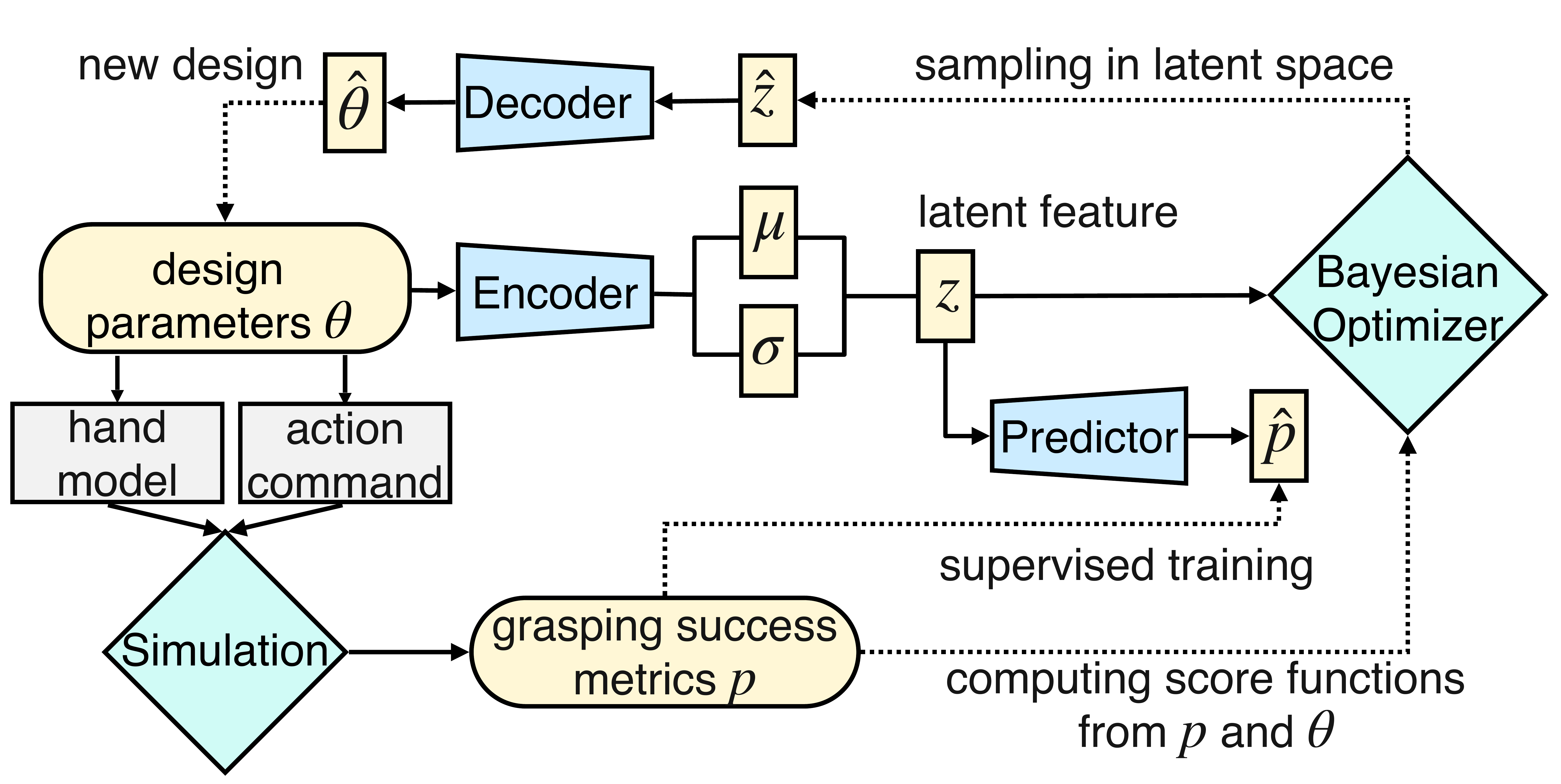}
\caption{\textbf{Schematic diagram of our  LABO model}. We learn a representation learning module, consisting of the encoder, decoder, and predictor to cast raw morphology and control parameters $\theta$ into a latent representation $z$. Then we perform Bayesian optimization in the latent space. We rely on simulated grasp trials to evaluate the generated hand morphology and control on success metrics $p$. 
}
\label{fig:rep}
\vspace{-10pt}
\end{figure}

 The co-optimization of morphology and control imposes significant challenges due to the high-dimensional parameter space, where na\"{i}ve sampling methods fail to discover effective designs. Our intuition is to exploit the correlation and redundancy in the design parameters -- some parameters have a stronger influence in task performance than others, and very often several of them work in synergy to shape the robot behaviors. For this reason, learning latent representations of the design parameters could capture such relationships among the parameters, leading to efficient search in the design space.
 Therefore, we use a representation learning module that transforms the raw parameter space into a reduced latent space. The learned latent space can be seamlessly integrated with distributed Bayesian optimization algorithms for black-box optimization of grasp quality metrics. 
 We name our proposed approach LAtent-space Bayesian Optimization (LABO), as shown in Figure~\ref{fig:rep}. 
Next, we first give a formal problem formulation and then introduce our LABO model.

\subsection{Problem Formulation}
We define the morphology and control parameters as $\theta$, where $\theta\in\mathbb{R}^d_{\geq0}$ and $d$ is the total dimension of the parameter space. A parameter $\theta$ can be further partitioned into two components $(\theta_m, \theta_a)$, where $\theta_m$ specifies the robot hand model, such as the number of fingers and their lengths, and $\theta_a$ defines the action command that actuates the hand for grasping, and we use torque control for each joint in the robot. Our goal is to maximize a black-box objective function $F$ by searching for a specific combination of morphology and control parameter $\theta^*$ such that
\begin{equation}
    \theta^* = \arg\max_{\theta}F(\theta),
\end{equation}
where $F$ is a score function that evaluates the quality of individual morphology and control designs. In practice, $F$ can be a simulator that instantiates the hand model and performs grasp trials to evaluate empirical performances of the design. Unlike prior work that used differentiable simulation~\cite{NIPS2019_9038,hu2019chainqueen}, we only assume black-box access to this function, and each call of the function incurs non-trivial computational cost, for instance, running physical simulation of grasping diverse objects.

We define the grasping task space as $\mathcal{T}$. 
Each unique task $t\in \mathcal{T}$ corresponds to a specific grasp type (power, pinch, or lateral) and a specific object instance. 
The task $t$ will take in the morphology and policy parameter $\theta$ and return the episodic reward obtained: $\mathcal{R}(\theta)$. The reward function is related with the quality and success rate of the grasping, also related with how robust it is against perturbations. In addition, we want to encourage a cost-effective design of the hand, such that non-functioning hand parts will be suppressed. The cost is task independent, and each hand design corresponds to a cost value $C(\theta)$, which corresponds to the morphology complexity. The score function is thus a weighted combination of the rewards and the cost:
\begin{equation}
    F(\theta) = w_1C(\theta)+w_2\sum_{i=1}^{|\mathcal{T}|}\mathcal{R}_i(\theta),
    \label{eq:score-function}
\end{equation}
where $w_1$ is the weight for the cost, and $w_2$ is the weight for the episodic rewards on all tasks, and $|\mathcal{T}|$ is the number of all tasks.

\textbf{Reward and Cost Functions.} The reward per step is defined per grasping task. Denote finger self-collision as $c_{s}$ (the value is 0 or 1), and the proportion of finger tips that are in contact with the object as $r_{o}$, the difference between the initial object position and orientation and the current object position and orientation as $d_{pos}, d_{orn}$, we have the following rewards definition:
\begin{align}
    \mathcal{R}_{power} & = -0.01 c_s + 0.1 r_o - 0.1 d_{pos},\\
    \mathcal{R}_{pinch} & = -0.01 c_s + 0.1 r_o - 0.1 d_{pos},\\
    \mathcal{R}_{lateral} & = -0.01 c_s + 0.1 r_o - 0.05 d_{pos} - 0.05 d_{orn}.
\end{align}
Define the number of fingers as $n_{f}$ and the number of finger segments as $n_{s,i}$, for $i\in\{1,2,\cdots, n_f\}$, the cost function is
\begin{align}
C & =\frac{n_f-2}{4} + \sum_{i=1}^{n_f}\frac{n_{s,i}-3}{3}.
\end{align}
The grasp success is checked by applying external adversarial perturbation forces from random directions at the center of mass of the object. If the object is held tightly in hand without dropping, the grasp is deemed a success; otherwise, a failure. The action space is consisted of the joint torques applied on all finger joints.

\subsection{Representation Learning of Design Parameters}
Directly optimizing in the raw parameter space suffers from the curse of dimensionality. To reduce the problem complexity and encourage knowledge transfer among different hand designs, we develop an encoder-decoder architecture to 
learn a latent space of design parameters, and use auxiliary losses to facilitate the representations to capture task-oriented information.

Denote the encoder network as $\phi_{E}$ and decoder network as $\phi_{D}$. The encoder network casts raw parameters $\theta$ into a probabilistic lower-dimensional latent space characterized by the mean $\mu$ and the standard deviation $\sigma$, where we use the reparameterization trick~\cite{kingma2019introduction} to draw a sample in the latent space $z=\mu+\sigma \odot \epsilon$, where $\epsilon\sim\mathcal{N}(0,I)$. To ensure that the latent space representation is constricted within a bounded region for the ease of the subsequent optimization, we further take a sigmoid function on $z$: $f=\text{sigmoid}(z)=\frac{1}{1+e^{-z}}$ so that $f\in\mathbb{R}^{d'}_{\geq 0}$, where $d'\ll d$ and $\|f\|_{\infty}=1$. The decoder reconstructs the original input from the latent representation. This can be written as
\begin{equation}\small{
    \theta\xrightarrow{\phi_E}(\mu,\sigma)\xrightarrow{\epsilon\sim\mathcal{N}(0,1)}z=\mu+\sigma\odot\epsilon\rightarrow f=\text{sigmoid}(z)\xrightarrow{\phi_D}\hat{\theta}.}
\end{equation}
We further regularize the latent space by restricting the KL divergence between $\mathcal{N}(\mu,\sigma)$ and $\mathcal{N}(0,I)$. 
However, the simple auto-encoding objectives are insufficient to inform the representations about the quality of the parameters in terms of task performance. We thus propose to use the grasp success metrics $p$ as an additional supervision for learning the latent representation $f$, where $p$ is a binary vector of which each dimension corresponds to the grasp success label on each training object. We train another prediction network $\phi_{P}$ that predicts the values of $p$ from the representation $f$ with binary cross-entropy losses.
The information flow of the auxiliary task can be expressed as
\begin{equation}
    \theta\xrightarrow{\phi_E}z\rightarrow f\xrightarrow{\phi_P}\hat{p}.
\end{equation}
To sum up, the overall loss function for training the representation module is,
\begin{equation}
\begin{split}
    L_{rep}(\theta, p, \phi_E, \phi_D, \phi_P) = & \frac{1}{d}\|\theta-\hat{\theta}\|^2+CELoss(p,\hat{p})\\
    &+D_{KL}(\mathcal{N}(0,I)||\mathcal{N}(\mu,\sigma)),
    \label{eq:rep-loss}
\end{split}
\end{equation}
where $\hat{\theta}=\phi_D(f)$, $\hat{p}=\phi_P(f)$, $CELoss$ denotes the binary cross entropy loss, and $D_{KL}$ denotes the KL-divergence of two distributions. Note that, in our optimization, the reward function $\mathcal{R}(\theta)$ in Equation~\eqref{eq:score-function} is a dense reward function more informative than the binary grasp labels $p$.


\begin{algorithm}[t!]
    \caption{Algorithmic Framework for Morphology and Control Optimization (LABO)}\small{
    \begin{algorithmic}[1]
        \State {\bfseries Input:}
            \State \quad BO iterations: $N_{BO}$\;
            \State \quad Number of pretrain data points: $N_{pre}$\;
            \State \quad  pretrain iterations: $N_1$ \;
            \State  \quad score function (simulator) with input $\theta$: $F(\theta)$\;
            \State \quad UCB acquisition function hyperparameter: $\beta$ \;
            \State \quad representation learning iterations per BO iteration: $N_2$\;
            \State \quad number of samples to draw for each BO iteartion: $N_3$ \;
        \State {\bfseries Initialize:}\;
          \State \quad $\phi_E, \phi_D, \phi_P,A(\xi,\beta)$ \;
          \State \quad $\mathcal{D}=\{\}$  \algorithmiccomment{Pretraining dataset}\;
          \State \quad $Q=\{\}$ \algorithmiccomment{BO dataset}\;
        \State {\bfseries Start:}\;
          \State \quad $\mathcal{D}\sim Uniform[0,1]\in\mathbb{R}^{d}$ \algorithmiccomment{sample $N_{pre}$ pretraining data}\;
          \State \quad {\bfseries For} $i$ = 1 to $N_1$:\;
          \State \quad\quad update $\phi_E$ with $\nabla_{\phi_E}L_{pre}(\mathcal{D}, \phi_E, \phi_D)$\;
          \State \quad\quad update $\phi_D$ with $\nabla_{\phi_D}L_{pre}(\mathcal{D}, \phi_E, \phi_D)$\;
          
          \State \quad {\bfseries For} $i$ = 1 to $N_{BO}$:\;
          \State \quad\quad $f\sim A(\xi,\beta)$ and $\hat{\theta}=\phi_D(f)$ \algorithmiccomment{sample new data point}\;
          \State \quad\quad $q=(f, F(\hat{\theta}))$ \algorithmiccomment{evaluate the new data in simulation}\;
          \State \quad\quad $Q=Q\cup \{q\}$\;
          \State \quad\quad Fit $G(\xi)$ with $Q$ once every $N_3$ steps \algorithmiccomment{Equation~\eqref{eq:surrogate}}\;
          \State \quad\quad {\bfseries For} $j$ = 1 to $N_2$:\;
          \State \quad\quad\quad $\nabla_{\phi_E}=\nabla_{\phi_E}L_{rep}(Q,\phi_E,\phi_D,\phi_P)$\; \State \quad\quad\quad\quad\quad\quad\quad $+\nabla_{\phi_E}L_{pre}(\mathcal{D},\phi_E,\phi_D)$\;
         \State \quad\quad\quad update $\phi_E$ with $\nabla_{\phi_E}$\;
          \State \quad\quad\quad $\nabla_{\phi_D}=\nabla_{\phi_D}L_{rep}(Q,\phi_E,\phi_D,\phi_P)$\; \State \quad\quad\quad\quad\quad\quad\quad $+\nabla_{\phi_D}L_{pre}(\mathcal{D},\phi_E,\phi_D)$\;
          \State \quad\quad\quad update $\phi_D$ with $\nabla_{\phi_D}$\; 
          \State \quad\quad\quad 
          $\nabla_{\phi_P} = \nabla_{\phi_P}L_{rep}(Q,\phi_E,\phi_D,\phi_P)$\;
          \State \quad\quad\quad update $\phi_P$ with $\nabla_{\phi_P}$\;
    \end{algorithmic}}
    \label{alg:flow}
\end{algorithm}

\subsection{Latent Space Bayesian Optimization}
Bayesian Optimization (BO)~\cite{frazier2018tutorial} is a classical family of methods to solve black-box optimization problems. Here we assume the function to be optimized is an unknown (and often costly) function with only inquiry access.  Thus, solving the optimization problem with as few samples as possible is vital. This is particularly important for us, as evaluating a hand on a set of hundreds of tasks takes millions of simulation steps, and fabricating a new hand design in the real world can be even more costly. Therefore, LABO harnesses the learned latent representations to accelerate the Bayesian optimization process.

\textbf{Pretrain the Representation Learning Module.} Before the optimization process starts, we pretrain the latent space by drawing random design parameters and train the encoder and decoder with the VAE objective~\cite{kingma2019introduction}. The pretraining loss function can be expressed as
\begin{equation}
    L_{pre}(\theta,\phi_E,\phi_D) = \frac{1}{d}\|\theta-\hat{\theta}\|^2+D_{KL}(\mathcal{N}(0,I)||\mathcal{N}(\mu,\sigma)).
    \label{eq:loss-pretrain}
\end{equation}
As we have not tested these designs in simulated trials, we do not have the grasp labels $p$ for supervising the auxiliary losses as in Equation~\eqref{eq:rep-loss}. Nonetheless, this procedure yields a good initialization of the latent representations to be further fine-tuned over the course of optimization.



\textbf{Joint Bayesian Optimization with Representation Learning.} After the initial pretraining, we start to use BO to perform optimization in the latent space. As more effective hand designs are expected to be gradually discovered, we can in turn use these new samples to update the representation module for subsequent optimization steps. For each BO iteration, we perform multiple iterations of representation learning. To speed up the process, rather than using a sequential sampler, we employ the parallelized Bayesian Optimization framework BoTorch~\cite{balandat2019botorch} for distributed sampling in the latent space. We use a Gaussian Process to model the data distribution, consisting of the morphology and control parameters in the latent space and their corresponding score value $F(\theta)$: $Q =\{q_i = (f_i, F(\theta_i))\}_{i=1}^N$, where $N$ is the number of data points the algorithm has seen so far. Given this dataset, we fit a surrogate distribution model of the data, with parameters $\xi$:
\begin{equation}
    G(q_{1:k};\xi)\sim \text{Normal}(\mu(q_{1:k}), \Sigma(q_{1:k}, q_{1:k});\xi),
    \label{eq:surrogate}
\end{equation}
where $\Sigma(q_{1:k}, q_{1:k})$ is the kernel function and we use the Matern kernel
\begin{equation}
    \Sigma(x,x') = \alpha\frac{2^{1-v}}{\Gamma(v)}(\sqrt{2v}\|x-x'\|)^vK_v(\sqrt{2v}\|x-x'\|),
\end{equation}
where $K_v$ is a modified Bessel function with parameter $v$, and $\alpha$ is another hyperparameter. We use Upper Confidence Bound (UCB) as our acquisition function:
\begin{equation}
    A(f;Q,\xi,\beta) = \mu(f;Q,\xi)-\beta\Sigma(f;Q,\xi),
\label{eq:acq}
\end{equation}
where $\beta$ is a parameter that can be tuned to trade off exploitation and exploration. The UCB algorithm is chosen since it has an additional parameter to tune how much exploration is done, which is important given the dimension of this optimization. We choose the Matern kernel since the changes in the morphology space is non-smooth (changes of numbers of fingers and finger segments), and a traditional Gaussian kernel might fail to model that well. Each time multiple data points are sampled to fit the surrogate model, and new samples are generated by maximizing the acquisition function. We include a complete pseudocode in Algorithm~\ref{alg:flow}.

\section{Experiments}
To operationalize our idea, we develop a new simulated platform in BulletPhysics~\cite{coumans2017pybullet} that supports procedural generation of parameterized robot hand morphologies and actuating the hand with joint-space controllers. We design our experiments to investigate the following questions: 1) Can the proposed model discover robust hand designs of high success rate and low cost, in comparison to baseline methods? 2) How well does the final hand design generalize to unseen objects in different grasp types? 3) How does the performance change with respect to objects of varying complexity? 4) How is the performance of simpler hand with fewer fingers compared to complex hand with more fingers. To answer these questions, we compare with several baselines and present both quantitative and qualitative results in this section. 

\subsection{Experiment Setup}

\begin{figure}[t!]
    \centering
    \begin{subfigure}[t]{0.45\linewidth}
        \centering
        \includegraphics[height=0.9in]{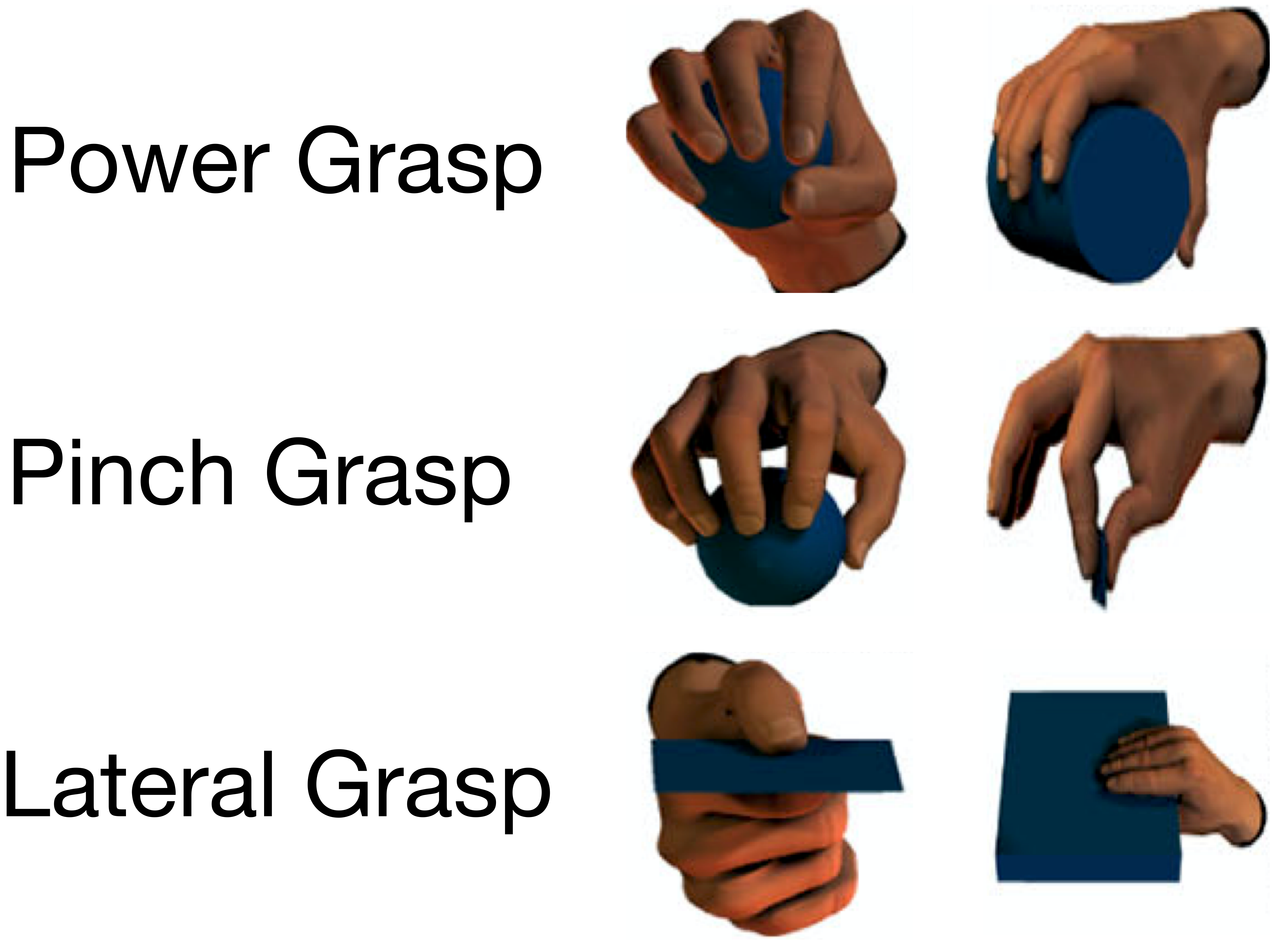}
        \caption{}
        \label{fig:grasptype}
    \end{subfigure}
    \begin{subfigure}[t]{0.53\linewidth}
        \centering
        \includegraphics[height=0.9in]{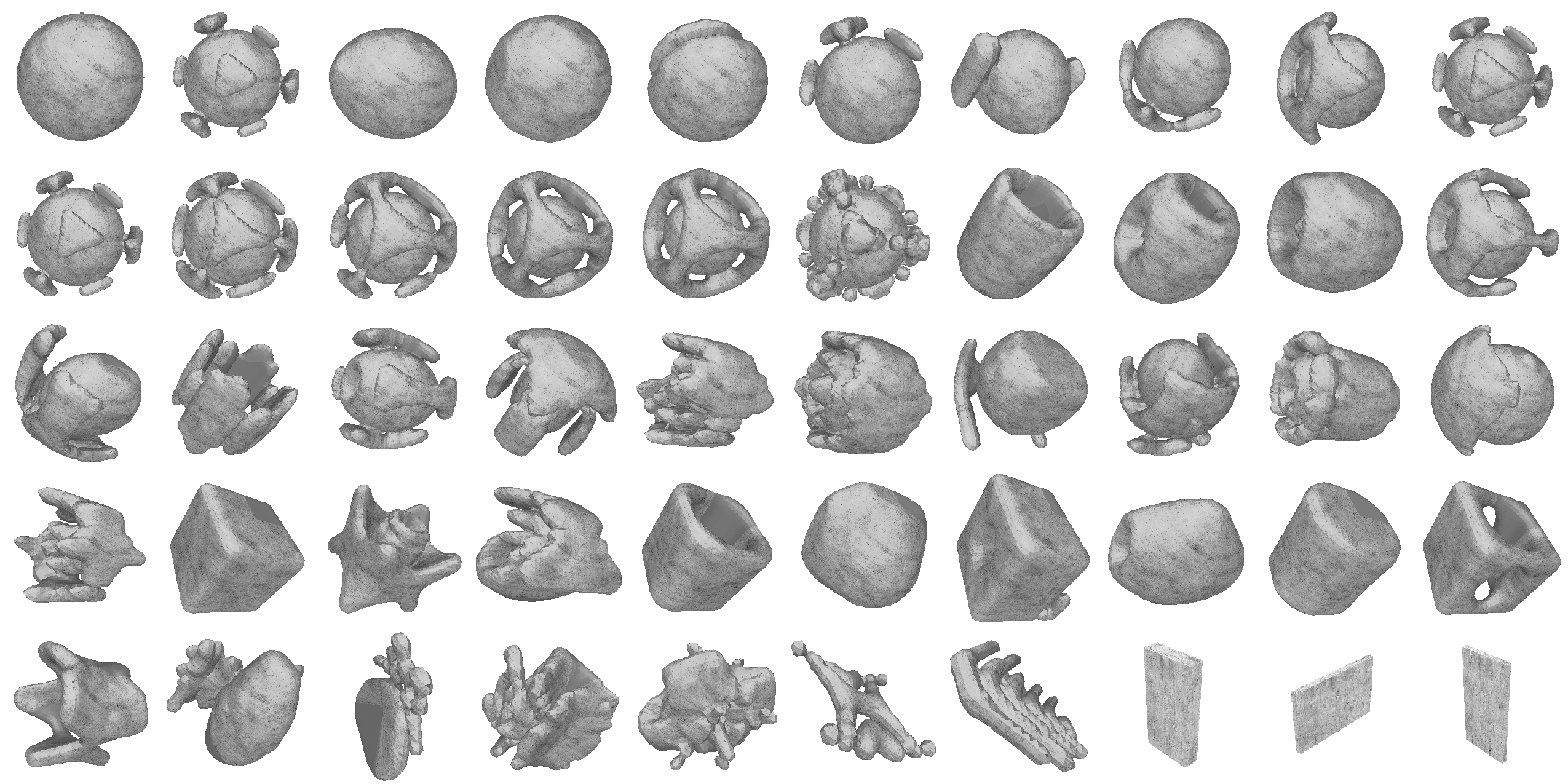}
        \caption{}
        \label{fig:object-dataset}
    \end{subfigure}
    \caption{(a) Three main grasp types of the human grasp taxonomy: power grasp, lateral grasp, and pinch grasp (images are taken from Feix et al.~\cite{feix2015grasp}); (b) Example objects we used for evaluating grasps. These objects are either selected from the EGAD dataset~\cite{morrison2020egad} or procedurally generated.}
    \vspace{-10pt}
\end{figure}

\textbf{Grasp Types.} To facilitate robust morphologies to emerge, we select three types of grasps from the human grasp taxonomy~\cite{feix2015grasp}, including power, pinch, and lateral grasp. We give definitions of these types of grasping tasks and provide a visualization of these types of grasps in Figure~\ref{fig:grasptype}:
\vspace{1mm}
\begin{enumerate}[leftmargin=*]
    \item \textit{Power Grasp}: The hand exerts forces to wrap the objects, with the surface of these fingers in a close contact with the object and the palm supporting the object. In this task, the palm faces upward and the object starts within the hand and we let the hand close to perform the grasp. 
    \item \textit{Pinch Grasp}: The fingers mainly rely on the fingertips to pick up the object, and the palm region will not support the object as well. In this task, we keep the palm facing downward and the hand will pick up an object below it and raise the object to a certain height.
    \item \textit{Lateral Grasp}: The hand uses finger inner surface and tips to support the object. Two set of fingers support opposite sides of an object. In this task, the target objects are mainly shaped like a thin plane. The hands will have the palm facing horizontally and try to grasp a plane object from the side of the object.
\end{enumerate}
\vspace{1mm}

\textbf{Object Datasets.} To get a diverse set of objects that demand different grasp strategies, we use the objects from the recently introduced EGAD dataset~\cite{morrison2020egad} and select 48 objects of varying grasp difficulties to assist evaluating the performance of the optimized hand on power and pinch grasps. For lateral grasps, we generate thin boxes with different sizes and shapes. Example objects are shown in Figure~\ref{fig:object-dataset}. The objects range from regular polygons, such as spheres and boxes, to irregular shapes, such as mugs.

\textbf{Parameterized Design Space.} Our original parameter space consists of 185 dimensions, with each dimension ranging between 0 and 1, denoted as $\theta\in\mathbb{R}^{185}_{\geq 0}$, where $\|\theta\|_{\infty}=1$. The first 122 dimensions correspond to the parameters $\theta_m$ that determine a specific hand design and the rest 63 dimensions determine the control $\theta_a$ for the three types of grasping tasks. For simplicity, we use open-loop control and share the same action command for objects in each grasp type. The 122 dimensions include parameters for number of fingers, mass, surface friction, finger shape parameters. The additional dimensions that determines the policy of the robot are related with the torque applied on each of the joints. These parameters are chosen since they affect the structure and the physical characteristics of the hand. The control policy parameters have 63 dimensions, where each grasping task (power, pinch, lateral) get 21 dimensions. These 21 dimensions specified the torque applied on the palm-finger joints, finger-self joints and finger-tip joints for a maximum of 6 fingers and 6 finger segments. 

\textbf{Comparative Study of Hand Complexity.} To analyze how grasping performance varies with respect to the hand complexity, we compare the performance of hands with different numbers of fingers. We fix the number of fingers of the hand to be 2 to 6, and perform optimization with LABO.

\subsection{Experiment Results}

\textbf{Baseline Comparisons}. We compare LABO with three baselines for black-box optimization:
\vspace{1mm}
\begin{enumerate}[leftmargin=*]
    \item \textit{Uniform Sampling} (Uniform): we randomly sample morphology and control parameters uniformly from [0,1] in the original raw parameter space and run the experiments for multiple trials and select the best design with the highest overall success rate as the final design.
    \item \textit{CMA-ES}: We use the covariance matrix adaptation evolutionary strategy (CMA-ES)~\cite{hansen2003reducing} approach and select the design of the highest overall success rate.
    \item \textit{BO w/ Raw Parameters} (BO): We use the same BO algorithm as our final model, but optimization runs in the original parameter space as opposed to the learned latent space.
\end{enumerate}
\vspace{1mm}
Table~\ref{tab:success_rate} reports the per-task and overall success rate. In terms of the final optimized hand cost values, they are $2.81\pm 1.12 $, $2.88\pm0.19$, $2.63\pm0.14$, and $1.83 \pm 0.67$ for Uniform, CMA-ES, BO and LABO, respectively. For all the methods, we sample the same number of data points. The results show that LABO outperforms other baselines on all three types of grasps with the lowest cost of the final hand design. 
The closest baseline is BO w/ Raw Parameters.
On the other hand, the Uniform Sampling and CMA-ES baselines fail to produce robust hand designs, resulting in more than 20\% lower success rate than ours and increased costs. Qualitatively speaking, the hand with a higher cost usually have redundant finger segments that are not functioning. The baselines can still find some successful hand designs but fail to remove these non-functioning fingers or segments due to the high dimensionality of the optimization problem. Optimizing in the latent space using LABO effectively find the important factors for design and can find more cost-effective hands.



\begin{figure}[t!]
    \centering
    \begin{subfigure}[t]{.49\linewidth}
        \centering
        \includegraphics[width=1.\linewidth]{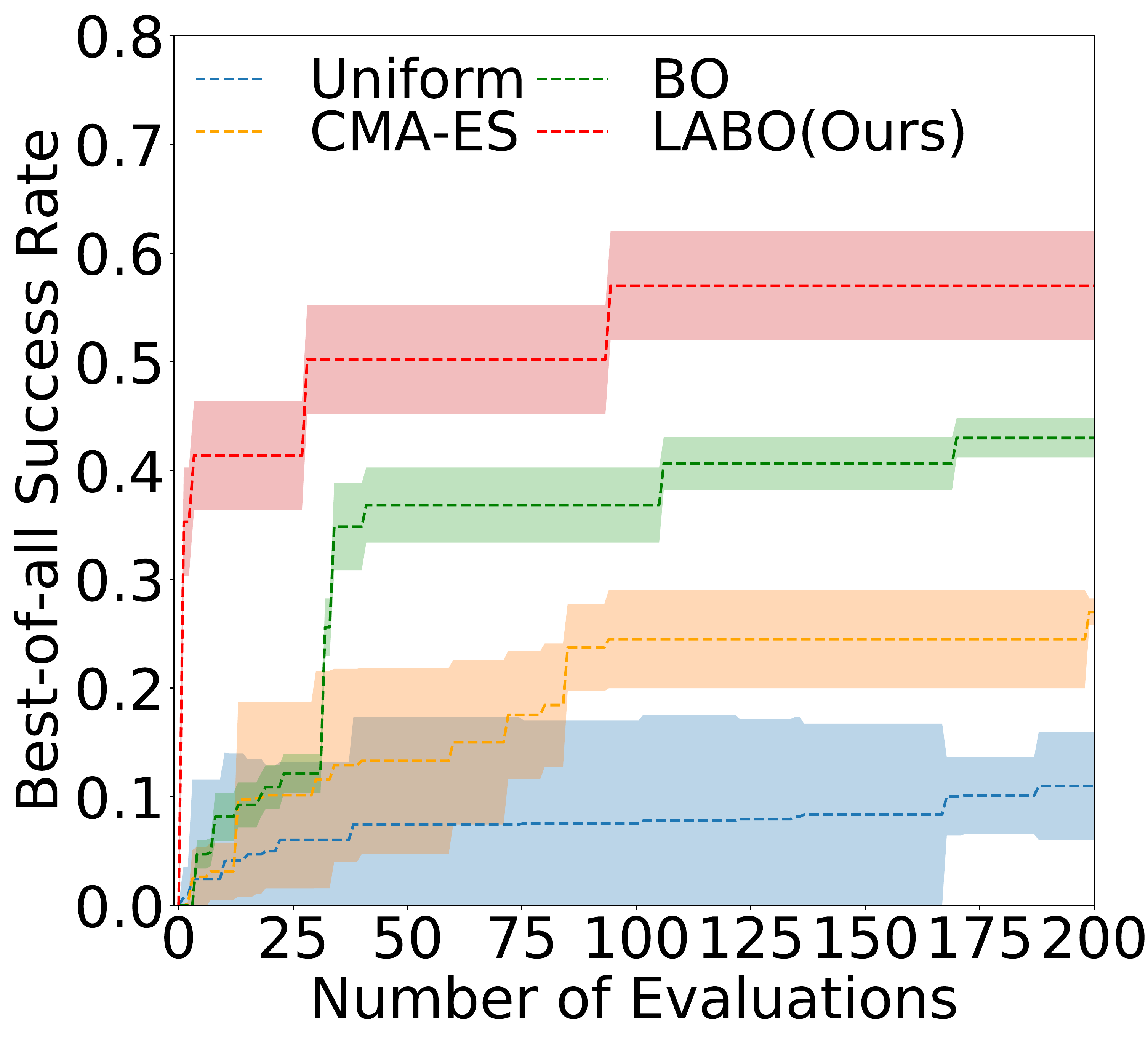}
        \caption{}
        \label{fig:training-curve}
    \end{subfigure}
    \begin{subfigure}[t]{.48\linewidth}
       \centering
       \includegraphics[width=1.\linewidth]{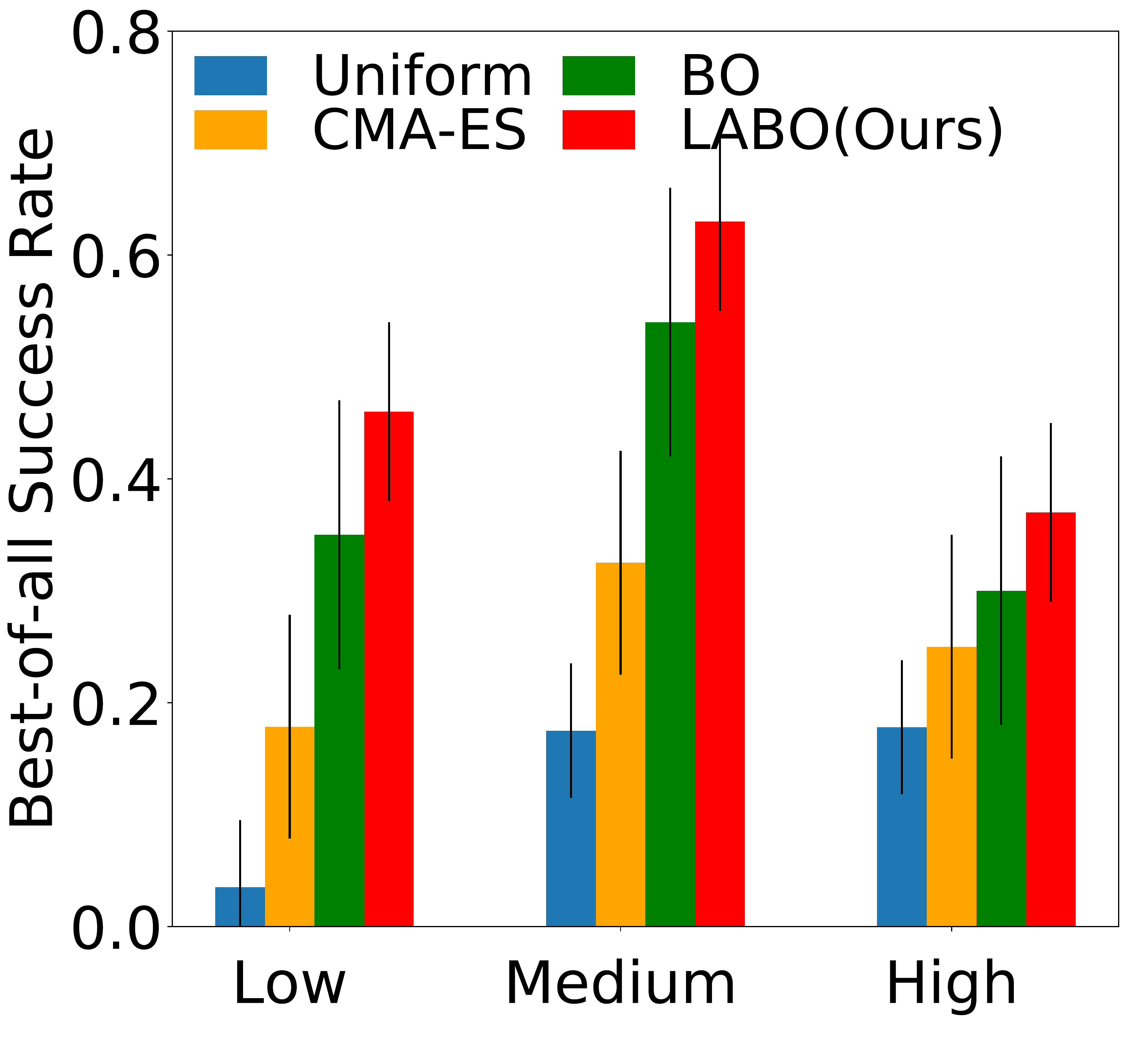}
       \caption{}
       \label{fig:barchart}
    \end{subfigure}
    \caption{(a) Best-of-all grasping success rates of the optimal hand designs on the training data set of objects; (b) Grasping success rates on objects of different geometric complexities.}
    \vspace{-10pt}
\end{figure}

Figure~\ref{fig:training-curve} shows the success rate curves of the all methods over the course of optimization. We found that the our model not only achieved the highest final performance, but also outperformed all baselines with less number of evaluations, and to achieve a success rate of 40\%, it is around 100 iterations faster than BO with raw parameters. It implies that the learned latent representations facilitate efficient search of promising samples and knowledge sharing across different design parameters, reducing the sample complexity of the Bayesian Optimization process. We further analyze the impact of object geometry on the grasp performance, and show an ablation study of the success rates of our optimal hand on these tasks with respect to the object geometric complexity in Figure~\ref{fig:barchart}. The geometric complexity is measured by the number of vertices in the 3D mesh models of the objects, where object models of less than 5k vertices are classified as \textit{low} and object models of greater than 7k vertices \textit{high}, and \textit{medium} otherwise. The results show that objects of medium complexity are easiest to grasp. We hypothesize that objects of low complexity, such as spheres, might lack flat surface for stable grasps, and those of high complexity, such as flower shapes, might be difficult to reach contact points. 
Figure~\ref{fig:optimized-hand} shows the qualitative examples of the optimized hands from our model and the grasping behaviors on objects of the three grasp types. The optimized hand developed different strategies for grasping objects of various shapes from the three grasp types. With the morphological costs as part of the optimization objective, it constrained the complexity of the design and landed on a three-finger hand.

\paragraph{Ablations on the Hand Complexity} In Table~\ref{tab:number_of_finger}, we show the result of the per-task grasping success rate, overall success rate for the optimized hand by fixing the number of fingers. It shows that simple hand with only 2 fingers can hardly succeed at the grasping tasks. We hypothesize that two-fingers hand is hard to form a tight force-closure and these hands are vulnerable to adversarial perturbations. With three or four fingers, the hand are becoming more robust to external perturbations while still keep the morphology cost-effective. Increased number of fingers cause a significantly increased overall success rate variance. This might be that with more fingers to design, there are more choices to make and less easy to make consistent designs. 

\begin{table}[t!]
    \scriptsize{
    \centering
    \caption{Success Rate of Optimized Hands on Novel Objects}
    \begin{tabular}{l|c|c|c|c}
        \toprule
       Methods & Power & Pinch & Lateral & Overall \\
        \midrule
        Uniform & 0.13 $\pm$ 0.08 & 0.03 $\pm$ 0.05 & 0.08 $\pm$ 0.05 & 0.07 $\pm$ 0.05 \\
        CMA-ES~\cite{hansen2003reducing} & 0.19 $\pm$ 0.20 & 0.14 $\pm$ 0.09 & 0.45 $\pm$ 0.40 & 0.26 $\pm$ 0.03 \\
        BO~\cite{frazier2018tutorial} & 0.24 $\pm$ 0.12 & 0.15 $\pm$ 0.14 & 0.72 $\pm$ 0.31 & 0.37 $\pm$ 0.19 \\
        LABO & \textbf{0.37 $\pm$ 0.19} & \textbf{0.21 $\pm$ 0.15} & \textbf{0.92 $\pm$ 0.10} & \textbf{0.50 $\pm$ 0.12}\\ 
        \bottomrule
    \end{tabular}
    \label{tab:success_rate}
    \vspace{-5pt}
    }
\end{table}

\begin{table}[t!]
    \centering
    \caption{Ablations Different Numbers of Fingers (Success Rate)}
    \scriptsize{
    \begin{tabular}{l|c|c|c|c}
    \toprule
        \# Fingers & Power & Pinch & Lateral  & Overall \\
        \midrule
         2 & 0.07~$\pm$~0.04 & 0.01~$\pm$~0.01 & 0.53~$\pm$~0.02 & 0.20~$\pm$~0.01\\
         3 & 0.37~$\pm$~0.07 & 0.24~$\pm$~0.02 & \textbf{0.97~$\pm$~0.06} & \textbf{0.52~$\pm$~0.04} \\
         4 & \textbf{0.38~$\pm$~0.24} & 0.22~$\pm$~0.18 & 0.71~$\pm$~0.27 & 0.44~$\pm$~0.09\\
         5 & 0.33~$\pm$~0.19 & 0.20~$\pm$~0.14 &  0.79~$\pm$~0.35 & 0.44~$\pm$~0.17\\
         6 & 0.31~$\pm$~0.24 & \textbf{0.25~$\pm$~0.15} & 0.95~$\pm$~0.07 & 0.50~$\pm$~0.13 \\
        \bottomrule
    \end{tabular}
    \label{tab:number_of_finger}
    \vspace{-10pt}
    }
\end{table}

\section{Conclusion}
In this work, we explore the emergence of embodiment and behavior of a robot hand to fulfill grasping tasks. Our goal is to discover robust and cost-efficient hand morphologies capable of grasping a diverse set of objects. To combat the challenge of searching in the high-dimensional design space, we integrate the learning of latent representations with scalable Bayesian optimization algorithms.
It validates our hypothesis that morphology and control can arise as emergent phenomena for the purpose of performing tasks. While our optimization method is general, two notable limitations could hinder its real-world use for designing robot hands. First, the optimization objective does not consider all practical constraints and costs of mechanical design. Our model simplified important factors, such as  actuation limits, fabrication costs, and energy efficiency, when evaluating the design samples. Second, using a physical simulator for grasp trials might result in vastly different outcomes than real-world experiments. A careful quantification of the simulation quality is necessary to ensure sim-to-real transfer of the grasp behaviors. For future work, we would like to explore the learning of vision-based closed-loop grasp policies in replacement of the open-loop controller. We also hope to take into account hardware constraints and the reality gap in order to extend our method to real-world robot designs.

\section*{ACKNOWLEDGMENT}
We gratefully acknowledge the feedback from members in NVIDIA AI Algorithms research team. We also acknowledge Jonathan Tremblay from NVIDIA for the support on ViSII rendering.

\bibliographystyle{IEEEtran}
\bibliography{example}  

\begin{thebibliography}{10}
\providecommand{\url}[1]{#1}
\csname url@rmstyle\endcsname
\providecommand{\newblock}{\relax}
\providecommand{\bibinfo}[2]{#2}
\providecommand\BIBentrySTDinterwordspacing{\spaceskip=0pt\relax}
\providecommand\BIBentryALTinterwordstretchfactor{4}
\providecommand\BIBentryALTinterwordspacing{\spaceskip=\fontdimen2\font plus
\BIBentryALTinterwordstretchfactor\fontdimen3\font minus
  \fontdimen4\font\relax}
\providecommand\BIBforeignlanguage[2]{{%
\expandafter\ifx\csname l@#1\endcsname\relax
\typeout{** WARNING: IEEEtran.bst: No hyphenation pattern has been}%
\typeout{** loaded for the language `#1'. Using the pattern for}%
\typeout{** the default language instead.}%
\else
\language=\csname l@#1\endcsname
\fi
#2}}

\bibitem{mautner2000evolving}
C.~Mautner and R.~K. Belew, ``Evolving robot morphology and control,''
  \emph{Artificial Life and Robotics}, vol.~4, no.~3, pp. 130--136, 2000.

\bibitem{young2003evolution}
R.~W. Young, ``Evolution of the human hand: the role of throwing and
  clubbing,'' \emph{Journal of Anatomy}, vol. 202, no.~1, pp. 165--174, 2003.

\bibitem{mahler2017dex}
J.~Mahler, M.~Matl, X.~Liu, A.~Li, D.~Gealy, and K.~Goldberg, ``Dex-net 3.0:
  Computing robust robot vacuum suction grasp targets in point clouds using a
  new analytic model and deep learning,'' \emph{arXiv preprint
  arXiv:1709.06670}, 2017.

\bibitem{mahler2017dex2}
J.~Mahler, J.~Liang, S.~Niyaz, M.~Laskey, R.~Doan, X.~Liu, J.~A. Ojea, and
  K.~Goldberg, ``Dex-net 2.0: Deep learning to plan robust grasps with
  synthetic point clouds and analytic grasp metrics,'' \emph{arXiv preprint
  arXiv:1703.09312}, 2017.

\bibitem{Shao2019UniGraspLA}
L.~Shao, F.~Ferreira, M.~Jorda, V.~Nambiar, J.~Luo, E.~Solowjow, J.~A. Ojea,
  O.~Khatib, and J.~Bohg, ``Unigrasp: Learning a unified model to grasp with
  n-fingered robotic hands,'' \emph{ArXiv}, vol. abs/1910.10900, 2019.

\bibitem{pfeifer2009morphological}
R.~Pfeifer and G.~G{\'o}mez, ``Morphological computation--connecting brain,
  body, and environment,'' in \emph{Creating brain-like intelligence}.\hskip
  1em plus 0.5em minus 0.4em\relax Springer, 2009, pp. 66--83.

\bibitem{backus2016adaptive}
S.~B. Backus and A.~M. Dollar, ``An adaptive three-fingered prismatic gripper
  with passive rotational joints,'' \emph{IEEE Robotics and Automation
  Letters}, vol.~1, no.~2, pp. 668--675, 2016.

\bibitem{deimel2016novel}
R.~Deimel and O.~Brock, ``A novel type of compliant and underactuated robotic
  hand for dexterous grasping,'' \emph{The International Journal of Robotics
  Research}, vol.~35, no. 1-3, pp. 161--185, 2016.

\bibitem{eppner2017lessons}
C.~Eppner, S.~H{\"o}fer, R.~Jonschkowski, A.~Sieverling, V.~Wall, and O.~Brock,
  ``Lessons from the amazon picking challenge: Four aspects of building robotic
  systems,'' in \emph{h International Joint Conference on Artificial
  Intelligence}, 2017.

\bibitem{yuan2020design}
S.~Yuan, A.~D. Epps, J.~B. Nowak, and J.~K. Salisbury, ``Design of a
  roller-based dexterous hand for object grasping and within-hand
  manipulation,'' in \emph{International Conference on Robotics and Automation
  (ICRA)}.\hskip 1em plus 0.5em minus 0.4em\relax IEEE, 2020.

\bibitem{feix2015grasp}
T.~Feix, J.~Romero, H.-B. Schmiedmayer, A.~M. Dollar, and D.~Kragic, ``The
  {GRASP} taxonomy of human grasp types,'' \emph{IEEE Transactions on
  human-machine systems}, vol.~46, no.~1, pp. 66--77, 2015.

\bibitem{hansen2003reducing}
N.~Hansen, S.~D. M{\"u}ller, and P.~Koumoutsakos, ``Reducing the time
  complexity of the derandomized evolution strategy with covariance matrix
  adaptation (cma-es),'' \emph{Evolutionary computation}, vol.~11, no.~1, pp.
  1--18, 2003.

\bibitem{frazier2018tutorial}
P.~I. Frazier, ``A tutorial on bayesian optimization,'' \emph{arXiv preprint
  arXiv:1807.02811}, 2018.

\bibitem{seok2014design}
S.~Seok, A.~Wang, M.~Y. Chuah, D.~J. Hyun, J.~Lee, D.~M. Otten, J.~H. Lang, and
  S.~Kim, ``Design principles for energy-efficient legged locomotion and
  implementation on the mit cheetah robot,'' \emph{Ieee/asme transactions on
  mechatronics}, vol.~20, no.~3, pp. 1117--1129, 2014.

\bibitem{graichen2015control}
K.~Graichen, S.~Hentzelt, A.~Hildebrandt, N.~K{\"a}rcher, N.~Gai{\ss}ert, and
  E.~Knubben, ``Control design for a bionic kangaroo,'' \emph{Control
  Engineering Practice}, vol.~42, pp. 106--117, 2015.

\bibitem{jayaram2020scaling}
K.~Jayaram, J.~Shum, S.~Castellanos, E.~F. Helbling, and R.~J. Wood, ``Scaling
  down an insect-size microrobot, hamr-vi into hamr-jr,'' \emph{arXiv preprint
  arXiv:2003.03337}, 2020.

\bibitem{leger1999automated}
C.~Leger \emph{et~al.}, \emph{Automated synthesis and optimization of robot
  configurations: an evolutionary approach}, 1999.

\bibitem{cheney2013unshackling}
N.~Cheney, R.~MacCurdy, J.~Clune, and H.~Lipson, ``Unshackling evolution:
  evolving soft robots with multiple materials and a powerful generative
  encoding,'' in \emph{Proceedings of the 15th annual conference on Genetic and
  evolutionary computation}, 2013, pp. 167--174.

\bibitem{luck2020data}
K.~S. Luck, H.~B. Amor, and R.~Calandra, ``Data-efficient co-adaptation of
  morphology and behaviour with deep reinforcement learning,'' in
  \emph{Conference on Robot Learning}, 2020, pp. 854--869.

\bibitem{wampler2009optimal}
K.~Wampler and Z.~Popovi{\'c}, ``Optimal gait and form for animal locomotion,''
  \emph{ACM Transactions on Graphics (TOG)}, vol.~28, no.~3, pp. 1--8, 2009.

\bibitem{ha2017joint}
S.~Ha, S.~Coros, A.~Alspach, J.~Kim, and K.~Yamane, ``Joint optimization of
  robot design and motion parameters using the implicit function theorem,'' in
  \emph{RSS}, 2017.

\bibitem{NIPS2019_9038}
A.~Spielberg, A.~Zhao, Y.~Hu, T.~Du, W.~Matusik, and D.~Rus,
  ``Learning-in-the-loop optimization: End-to-end control and co-design of soft
  robots through learned deep latent representations,'' in \emph{Advances in
  Neural Information Processing Systems 32}, 2019.

\bibitem{hu2019chainqueen}
Y.~Hu, J.~Liu, A.~Spielberg, J.~B. Tenenbaum, W.~T. Freeman, J.~Wu, D.~Rus, and
  W.~Matusik, ``Chainqueen: A real-time differentiable physical simulator for
  soft robotics,'' in \emph{International Conference on Robotics and Automation
  (ICRA)}, 2019.

\bibitem{chen2020hardware}
T.~Chen, Z.~He, and M.~Ciocarlie, ``Hardware as policy: Mechanical and
  computational co-optimization using deep reinforcement learning,''
  \emph{arXiv preprint arXiv:2008.04460}, 2020.

\bibitem{wang2019neural}
T.~Wang, Y.~Zhou, S.~Fidler, and J.~Ba, ``Neural graph evolution: Towards
  efficient automatic robot design,'' \emph{arXiv preprint arXiv:1906.05370},
  2019.

\bibitem{pathak2019learning}
D.~Pathak, C.~Lu, T.~Darrell, P.~Isola, and A.~A. Efros, ``Learning to control
  self-assembling morphologies: a study of generalization via modularity,'' in
  \emph{Advances in Neural Information Processing Systems}, 2019, pp.
  2295--2305.

\bibitem{chen2018hardware}
T.~Chen, A.~Murali, and A.~Gupta, ``Hardware conditioned policies for
  multi-robot transfer learning,'' in \emph{Advances in Neural Information
  Processing Systems}, 2018, pp. 9355--9366.

\bibitem{huang2020smp}
W.~Huang, I.~Mordatch, and D.~Pathak, ``One policy to control them all: Shared
  modular policies for agent-agnostic control,'' in \emph{ICML}, 2020.

\bibitem{bohg2013data}
J.~Bohg, A.~Morales, T.~Asfour, and D.~Kragic, ``Data-driven grasp
  synthesis—a survey,'' \emph{IEEE Transactions on Robotics}, vol.~30, no.~2,
  pp. 289--309, 2013.

\bibitem{lenz2015deep}
I.~Lenz, H.~Lee, and A.~Saxena, ``Deep learning for detecting robotic grasps,''
  \emph{The International Journal of Robotics Research}, vol.~34, no. 4-5, pp.
  705--724, 2015.

\bibitem{levine2018learning}
S.~Levine, P.~Pastor, A.~Krizhevsky, J.~Ibarz, and D.~Quillen, ``Learning
  hand-eye coordination for robotic grasping with deep learning and large-scale
  data collection,'' \emph{The International Journal of Robotics Research},
  vol.~37, no. 4-5, pp. 421--436, 2018.

\bibitem{pinto2016supersizing}
L.~Pinto and A.~Gupta, ``Supersizing self-supervision: Learning to grasp from
  50k tries and 700 robot hours,'' in \emph{IEEE International Conference on
  Robotics and Automation (ICRA)}, 2016.

\bibitem{fang2018tog}
K.~Fang, Y.~Zhu, A.~Garg, A.~Kuryenkov, V.~Mehta, L.~Fei-Fei, and S.~Savarese,
  ``Learning task-oriented grasping for tool manipulation from simulated
  self-supervision,'' \emph{Robotics: Science and Systems (RSS)}, 2018.

\bibitem{bousmalis2018using}
K.~Bousmalis, A.~Irpan, P.~Wohlhart, Y.~Bai, M.~Kelcey, M.~Kalakrishnan,
  L.~Downs, J.~Ibarz, P.~Pastor, K.~Konolige, \emph{et~al.}, ``Using simulation
  and domain adaptation to improve efficiency of deep robotic grasping,'' in
  \emph{IEEE international conference on robotics and automation (ICRA)}.\hskip
  1em plus 0.5em minus 0.4em\relax IEEE, 2018, pp. 4243--4250.

\bibitem{viereck2017learning}
U.~Viereck, A.~t. Pas, K.~Saenko, and R.~Platt, ``Learning a visuomotor
  controller for real world robotic grasping using simulated depth images,''
  \emph{arXiv preprint arXiv:1706.04652}, 2017.

\bibitem{andrychowicz2020learning}
O.~M. Andrychowicz, B.~Baker, M.~Chociej, R.~Jozefowicz, B.~McGrew,
  J.~Pachocki, A.~Petron, M.~Plappert, G.~Powell, A.~Ray, \emph{et~al.},
  ``Learning dexterous in-hand manipulation,'' \emph{The International Journal
  of Robotics Research}, vol.~39, no.~1, pp. 3--20, 2020.

\bibitem{gupta2016learning}
A.~Gupta, C.~Eppner, S.~Levine, and P.~Abbeel, ``Learning dexterous
  manipulation for a soft robotic hand from human demonstrations,'' in
  \emph{2016 IEEE/RSJ International Conference on Intelligent Robots and
  Systems (IROS)}.\hskip 1em plus 0.5em minus 0.4em\relax IEEE, 2016, pp.
  3786--3793.

\bibitem{dhamala2018high}
J.~Dhamala, S.~Ghimire, J.~L. Sapp, B.~M. Hor{\'a}{\v{c}}ek, and L.~Wang,
  ``High-dimensional bayesian optimization of personalized cardiac model
  parameters via an embedded generative model,'' in \emph{International
  Conference on Medical Image Computing and Computer-Assisted
  Intervention}.\hskip 1em plus 0.5em minus 0.4em\relax Springer, 2018, pp.
  499--507.

\bibitem{griffiths2017constrained}
R.-R. Griffiths and J.~M. Hern{\'a}ndez-Lobato, ``Constrained bayesian
  optimization for automatic chemical design,'' \emph{arXiv preprint
  arXiv:1709.05501}, 2017.

\bibitem{napier1962evolution}
J.~Napier, ``The evolution of the hand,'' \emph{Scientific American}, 1962.

\bibitem{van2009optimal}
E.~Van~Henten, D.~Van’t~Slot, C.~Hol, and L.~Van~Willigenburg, ``Optimal
  manipulator design for a cucumber harvesting robot,'' \emph{Computers and
  electronics in agriculture}, vol.~65, no.~2, pp. 247--257, 2009.

\bibitem{kingma2019introduction}
D.~P. Kingma and M.~Welling, ``An introduction to variational autoencoders,''
  \emph{arXiv preprint arXiv:1906.02691}, 2019.

\bibitem{balandat2019botorch}
M.~Balandat, B.~Karrer, D.~R. Jiang, S.~Daulton, B.~Letham, A.~G. Wilson, and
  E.~Bakshy, ``Botorch: Programmable bayesian optimization in pytorch,''
  \emph{arXiv preprint arXiv:1910.06403}, 2019.

\bibitem{coumans2017pybullet}
E.~Coumans and Y.~Bai, ``Pybullet, a python module for physics simulation in
  robotics, games and machine learning,'' 2017.

\bibitem{morrison2020egad}
D.~Morrison, P.~Corke, and J.~Leitner, ``Egad! an evolved grasping analysis
  dataset for diversity and reproducibility in robotic manipulation,''
  \emph{IEEE Robotics and Automation Letters}, 2020.

\end{thebibliography}

\end{document}


\maketitle

\section{Experiment Details}

\paragraph{Morphology and Controller Design Parameter Specifics} As mentioned in the main text, we have 185 dimensions for the morphology and controller design, where 122 dimensions correspond to morphology design and 63 dimensions correspond to controller design. All parameters are initially sampled from [0,1], and then mapped onto the actual parameters. Among the 122 morphology design dimensions, there is 1 dimension for determining the height where the hand will be fixed relative to the ground, with the height between -1 and 1; 1 dimension for the mass of hand finger segments (between 1 and 5, included); 1 dimension for determining the number of fingers (between 2 and 6, included), and for each of the finger, we have at least 2 segments and at most 6 segments. We use a 6 dimension vector to determine the number of segments for each finger and use a 96 dimension vector to represent the all possible kinds of hand finger morphology, indicating the height and radius of each of the finger segments. The height of each segment is bounded between 1 and 1.5, and the radius of each segment is bounded between 0.2 and 0.4. In the case where there are less than 6 fingers, the parameters which have no corresponding finger or corresponding finger segments will still be assigned values but they are not used while we evaluate the performance of the hand. In addition, we use a 6 dimension vector for determining finger tip (which is a sphere) shape parameters, and a 5 dimension vector for determining the friction (lateral and spinning friction, between 1 and 5) and joint property of the hand (velocity limit between 0.1 and 2, damping between 1 and 1.1, and effort between 500 and 4000). Finally, we use a 6 dimension vector for determining the location where each of the finger will be mounted on the palm, which is a thin cylinder.

For the control parameters, we assign a target velocity to each of the movable joints, which will be specified with the 63 dimension vector, where each of 21 dimension vector will be responsible for encoding the control strategy for a type of grasping. We use open-loop control in this work. We have a reject mechanism to immediately reject a hand design when it is obvious that this morphology would not work: we bound the minimum angular distance between two fingers to by a lower bound 12 degrees and reject a sample that has two fingers too close. 

\paragraph{Detailed Reward Function} We mentioned in the main text of the reward and cost function we use in our experiment. We give the details of these functions here. For each task, the time gap for one step of simulation is set to be 1/240s, and we set a Pybullet physics engine number of solver iteration to be 200. One episode in each task runs for 2000 steps followed by 800 steps of random perturbations from 8 selected directions and random selected magnitude, where each direction random perturbation runs for 100 steps. The perturbations are directly applied on the object. The per step reward function is grasp type specific, and all involves penalizing self-collisions between finger segments and encouraging the contact of all finger tips with the object. We define the self-collision of the hand finger segments as the collision of non-adjacent finger segments. Define the indicator of self-collision as $I_{sc}$. We define the ratio of contact between finger tips and the object as $r_{to}$, which is the number of finger tips that is in contact with the object over the the number of total finger tips. We list the reward function for the three types of grasping here.

\begin{enumerate}
    \item \textit{Power Grasp}. Per step reward function when there is no perturbations is defined as $R=-0.01I_{sc}+0.01r_{to}$ . During the perturbation stage, we apply random perturbations from 8 selected directions and random selected magnitude (between 500 and 1000). Define an indicator $I_v$ indicating whether the object is within a vicinity of the initial position (instead of flying away or dropping out of the hand), and define the magnitude of the perturbation force as $m_f$, the per step reward function is defined as $R_p=I_v\times(m_f\times10^{-5})-0.05(1-I_v)+r_{to}\times10^{-5}$. Since the hand is facing upward, the vicinity check is to check whether in the x,y direction the object is within 3.0 units from the initial position and in the z direction the object is within 1.0 unit from the initial position.
    \item \textit{Pinch Grasp}. Using the same notation as in power grasp, the per step reward function when there is no perturbations is defined as $R=-0.01\times I_{sc}+0.01\times r_{to}$. The perturbation stage per step reward function is defined as $R_p=I_v\times(m_f\times10^{-5})-0.05(1-I_v)+r_{to}\times10^{-5}$. The vicinity check $I_v$ is to check whether in the z direction the object has been lifted for about at least 0.05 unit, and in the x,y direction the object should be within 2.0 from its initial position.
    \item \textit{Lateral Grasp}. Using the same notation as in power grasp, the per step reward function when there is no perturbations is defined as $R=-0.01\times I_{sc}+0.01\times r_{to}$. At the perturbation stage, we penalize the re-orientation of the object. Define the total absolute difference between the initial orientation of the object and the current orientation of the object as $d_{orn}$. The perturbation stage per step reward function is defined as $R_p=I_v\times(m_f\times10^{-5})-0.05(1-I_v)+r_{to}\times10^{-5}$.
    The vicinity check $I_v$ checks whether $d_{orn}$ is smaller than 2.5 and in the x,y direction the object should be within 3.0 unit from its initial position and in the z direction the object should be within 0.3 unit from its initial position. 
\end{enumerate}

\paragraph{Cost Function} The cost function is task-agnostic and only depends on the hand morphology. We limit the total number of fingers to be between 2 and 6 (included), and limit the number of finger segments to be between 3 and 6. Define the number of fingers to be $n_f$ and number of total finger segments to be $n_s$, the cost function is defined as,
\begin{equation}
    C(n_f, n_s) = \frac{n_f-2}{4} + \frac{n_s-3}{3}.
\end{equation}
The final score function of a morphology design and policy strategy depends on both the total reward in each task and the morphology cost. Define the number of total tasks as $n_t$, the reward for each task as $R_i$, with $i=1,\cdots, n_t$, the score function is defined as,
\begin{equation}
    F = \frac{1}{n_t}\sum_{i=1}^{n_t}R_i-0.1\times C(n_f,n_s).
\end{equation}

\paragraph{Success Check} The success check for each of the task is the vicinity check when there is perturbation in the environment to pull objects away from the hand. The success is indicated by the vicinity check indicator $I_v$ ($I_v=1$ means success). 

\section{Training Details}
For training with our approach using deep learning based Bayesian optimization, we use the open source environment Botorch~\cite{balandat2019botorch}. We use the default Matern kernel and single task Gaussian process to build the surrogate model. When optimizing the acquisition function for each step, we use 10 starting points for the multi-start acquisition function, and use 1024 raw samples for initialization, generating 2 candidates each time for the morphology evaluation function to evaluate. For training with representation learning, we use a simple multi-layer perceptron (MLP) for the encoder, decoder, and predictor. The encoder is a 4-layer network with 185 dimension input, 100 dimension hidden, and 32 dimension output. ReLU non-linear activation function is used between these linear layers. The decoder is a 3-layer network with 185 dimension output, 32 dimension input, and 100 dimension hidden. ReLU non-linear activation function is used between the hidden layers, and Sigmoid function is used in the output layer. The predictor is also a 3-layer MLP, with 160 dimension output, and 32 dimension input, and 100 dimension hidden. ReLU non-linear activation function is used between the hidden layers, and Sigmoid function is used in the output layer. For training the representation learning model, we use the Adam optimizer~\cite{kingma2014method} with a learning rate of 0.0001. We pretrain the VAE model until converge for 100,000 steps with a batch size of 32, and for each Bayesian optimization iteration, we further train the VAE model for 2000 steps, with a batch size of 32, where 1000 steps is used for training with the labeled data and 1000 steps is used for training with the pretraining data.

\section{Validity of Real World Applicability of the Proposed Work}
We did our evaluation of the hand design entirely in simulation. The simulator design already takes into consideration real physical constraints and limits, and similar experiments can be carried out in real environments. Since our approach does not require to learn a policy, once the hand parameters and controller parameters are selected, the hand can be directly manufactured and evaluated in the real world. Furthermore, we only used 200 samples and get a decent performance on a diverse set of objects, indicating that the same experiments can be done in the real world without the requirement of too much human labor.
On the other hand, the simulator can be improved to match with real world physical parameters and constraints more accurately, and there is a possibility to directly optimize in simulation and then transfer to real world applications. PyBullet is a very advanced simulation environment that is widely used and has been demonstrated to achieve sim-to-real transfer such as the work presented by~\cite{peng2020learning, yu2020learning}. Real world physical constraints can also be integrated with the simulator and our black-box optimizer will be able to cope with that. We leave this for the future work.

\clearpage
\bibliography{example}  

